%% file: main.tex
\documentclass[letterpaper, 10 pt, conference]{ieeeconf}  %
\IEEEoverridecommandlockouts                              
\usepackage{graphicx} 

\usepackage{amsmath} 
\usepackage{amssymb}  
\usepackage{booktabs}
\usepackage{cite}
\usepackage{lipsum}
\usepackage{comment}
\usepackage{flexisym}
\usepackage{amsmath}
\usepackage{subcaption}
\usepackage{flushend}
\usepackage{algpseudocode}
\usepackage{algorithm}

\algnewcommand\algorithmicforeach{\textbf{for each}}
\algdef{S}[FOR]{ForEach}[1]{\algorithmicforeach\ #1\ \algorithmicdo}

\newcommand{\ts}{\textsuperscript}

\usepackage{xcolor}


\title{\LARGE \bf Did You Miss the Sign? A False Negative Alarm System for Traffic Sign Detectors }
\author{Quazi Marufur Rahman and Niko S{\"u}nderhauf and Feras Dayoub %
\thanks{The authors are with the Australian Centre for Robotic Vision at Queensland University of Technology (QUT),
Brisbane, QLD 4001, Australia.  Contact: {\tt\small quazi.rahman@qut.edu.au}}%
}
\begin{document}

\maketitle
\begin{abstract}
Object detection is an integral part of an autonomous vehicle for its safety-critical and navigational purposes. Traffic signs as objects play a vital role in guiding such systems. However, if the vehicle fails to locate any critical sign, it might make a catastrophic failure. In this paper, we are proposing an approach to identify traffic signs that have been mistakenly discarded by the object detector. The proposed method raises an alarm when it discovers a failure by the object detector to detect a traffic sign. This approach can be useful to evaluate the performance of the detector during the deployment phase. We trained a single shot multi-box object detector to detect traffic signs and used its internal features to train a separate false negative detector (FND). During deployment, FND decides whether the traffic sign detector (TSD) has missed a sign or not. We are using precision and recall to measure the accuracy of FND in two different datasets. For $80\%$ recall, FND has achieved $89.9\%$ precision in Belgium Traffic Sign Detection dataset and $90.8\%$ precision in German Traffic Sign Recognition Benchmark dataset respectively. To the best of our knowledge, our method is the first to tackle this critical aspect of false negative detection in robotic vision. Such a fail-safe mechanism for object detection can improve the engagement of robotic vision systems in our daily life.

\end{abstract}

\section{Introduction}
Traffic sign detection and recognition are critical parts of an autonomous vehicle (AV) system for its navigational purpose. Currently, traffic sign detection is achieved by deploying state-of-the-art deep learning object detection networks~\cite{Wen2017traffic,zhu2016traffic,sheikh2016traffic,lee2018simultaneous}. For safety reasons, an AV should not miss detecting any sign as such a mistake could result in a catastrophic incident. Therefore, these detectors are required to operate reliably in variable conditions. However, unknown environments, degraded image quality due to bad weather, uneven illumination, and poor textures are some of the factors that can impact the performance of the deployed object detection systems onboard AVs. This fact raises safety concerns, and it is among the reasons that are halting the widespread deployment of autonomous vehicles beyond level~$2$ autonomy~\cite{sae2014taxonomy} where the vehicle must be assisted by the driver when needed \cite{litman2017autonomous}. 

One way to tackle the safety issue is to keep improving the performance of the traffic sign detectors. However and given the fact that object detection systems cannot be guaranteed never to make mistakes, we argue that there should be a mechanism to detect when these detectors make mistakes during deployment -- a failure detection system that raises the alarm when there is evidence that the performance of the deployed sign detector may have degraded. The aim is to alert the detector that it may have made a mistake to identify a sign in a particular region of its input image. Upon receiving the alarm, the detector can take alternative measures to detect the sign again. Consequently, if the frequency of alarm keeps increasing, the autonomous system can ask for intervention from a human user to take control.
\begin{figure}[tb]
    \centering
    \includegraphics[width=1.0\linewidth]{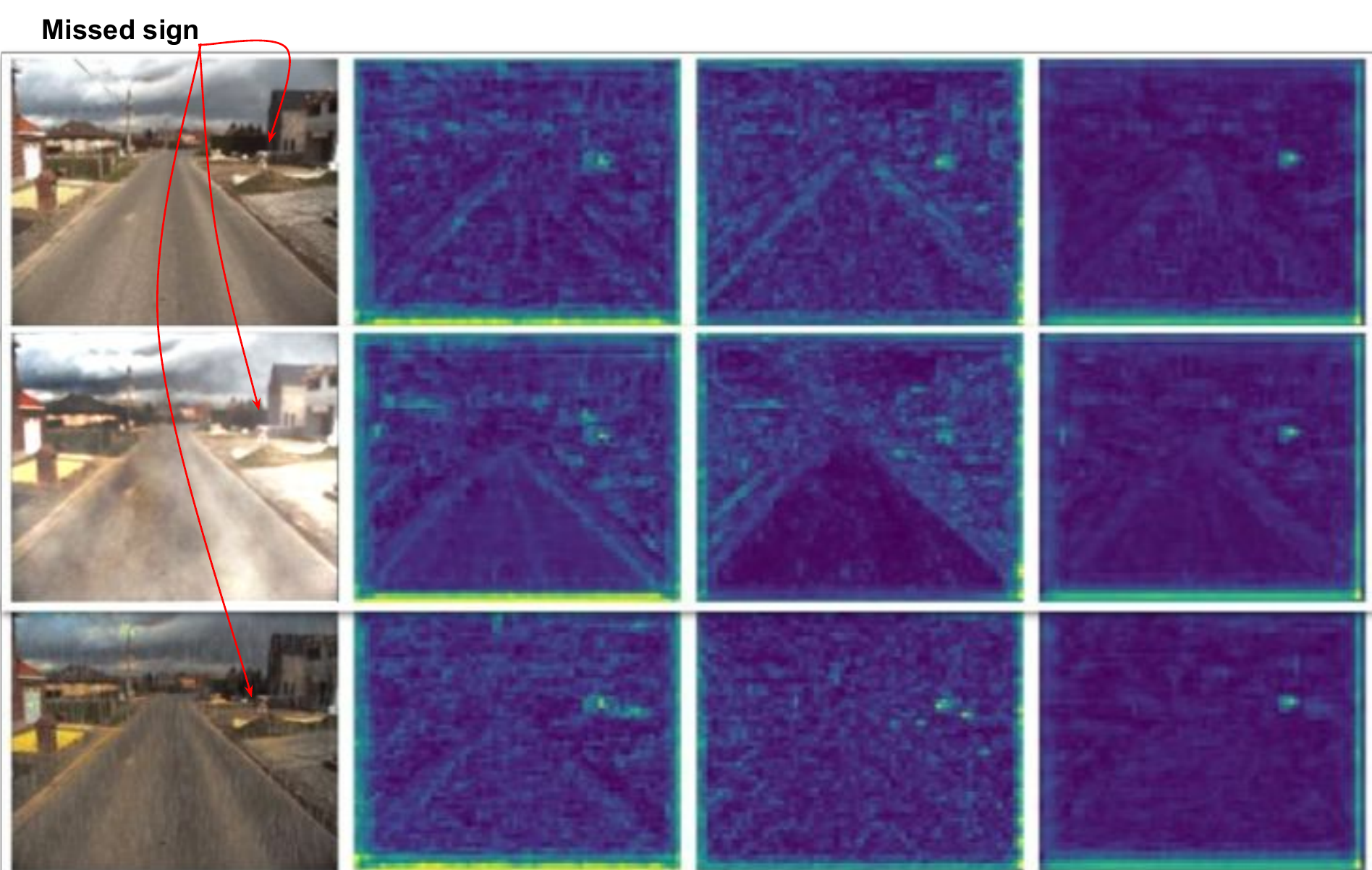}
    \caption{An example of input images with various weather conditions along with there feature maps. In all of these images, the deployed sign detector has failed to detect the traffic sign. However, the feature maps show excited regions some of them correspond to the missed traffic sign. During training of the failure detector, we extract these excited regions and label them as failures (false negative) or imposters (true negative). The three rows are showing the input image in normal, simulated foggy and simulated rainy weather conditions respectively.}
    \label{fig:main_idea}
\end{figure}

To this end, this paper proposes such a failure detection system in the context of traffic sign detection, although the proposed method is not restricted to traffic signs. Our proposed method uses the feature maps of a deployed traffic sign deep neural network to extract cues and detect potential false negatives. To the best of our knowledge, this method is the first to tackle this critical aspect of false negative detection in robotic vision. 

The rest of the paper is organized as follow: In Section~\ref{sec:related}, we review the related works on failure detection. In Section~\ref{sec:our_approach}, we introduce our approach to detect failure of a traffic sign detection system by discovering false negatives. Section~\ref{sec:experiments} outlines our experimental evaluation setup. Section~\ref{sec:eval} presents the results and finally in Section~\ref{sec:conclusions} we draw  conclusions  and suggest areas for future work.

\begin{figure*}[t]
    \centering
    \includegraphics[width=\textwidth]{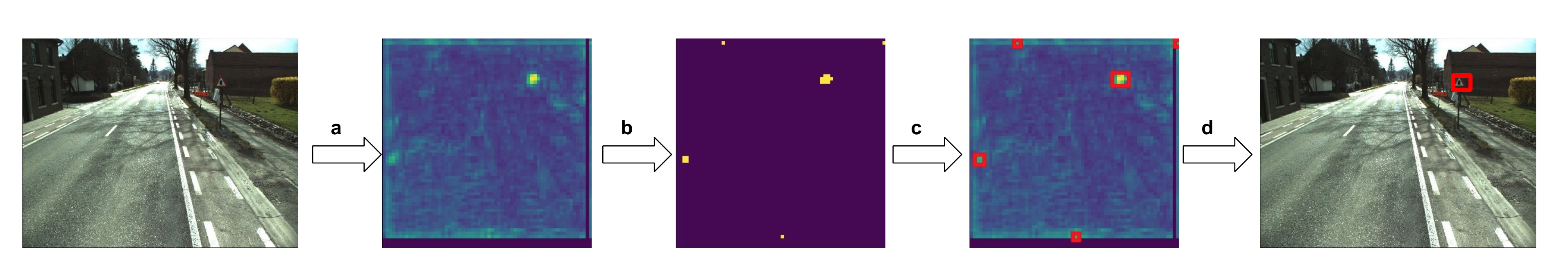}
    \caption{An illustration of the feature extraction pipeline. (a) Extraction of a 2D representation of a feature map generated by TSD for an input image where TSD has missed to detect the sign. FND produces this feature map by channel-wise max pooling operation. (b) Feature map is binarized to identify highly excited regions. (c) Contour area detection technique is applied to the binarized feature map to locate the excited areas. (d) Extracted features from the selected areas are used to detect false negative traffic signs.}
    
    \label{fig:examples}
\end{figure*}
\section{Related Work}\label{sec:related}

Several works have been done in the area of detecting or predicting failures of vision systems. We can categorize the proposed approaches into two broad groups. The first group identifies failures by examining the output of the vision system. The second group uses a separate system to predict the failure of the vision system based on its input. 

Among the first group of approaches, \cite{morris2007robotic} introduced the idea of introspection in the context of robotics. They described the ambiguity of a robots awareness during deployment as a barrier to using these systems in a real environment. \cite{grimmett2013knowing} and \cite{triebel2016driven} have discussed the importance of introspection capability in robotics context using a classifier. They have performed extensive studies to explore the failure prediction capability of multiple classification algorithms. They propose to assess the predictive variance to mitigate potentially overconfident classifiers. \cite{devries2018leveraging} proposed a single end-to-end framework to measure the uncertainty of an automated segmentation pipeline and \cite{devries2018learning} estimated the confidence of a neural network for out of distribution sample detection.

In the second group of procedures, \cite{zhang2014predicting} proposed a warning system named ALERT to build a self-evaluating vision system. It analyzes the input and predict the output reliability of the vision system. \cite{hecker2018failure} has proposed the concept of scene drivability. It predicts the feasibility of driving scene for a driving method. A probabilistic approach has been used by \cite{Gurau2018LearnFE} to use space, time and appearance to predict the performance of an autonomous vehicle. They also anticipate when to hand over control to the human user. \cite{hu2017introspective} proposed a method to evaluate the performance of a perception system without any ground truth. \cite{saxena2017learning} argued that most vision based perception failure occurs because of improper illumination of the scene. To tackle this problem, they proposed a failure detection and recovery maneuver for a vision system. A system agnostic framework has been proposed by \cite{daftry2016introspective} to predict failure in a vision system. They argued that predicting failure from raw sensor data is more effective than using the uncertainty of model-based classifiers.

Our proposed algorithm belongs to the first group of approaches as we use the traffic sign detector to extract important cues to discover the failure of the detector. As stated in the introduction, our work does not aim to improve the performance of the sign detector. Instead, we focus on identifying the cases where the detector fails to identify a traffic sign from a particular location.

\section{False Negative Detection}
\label{sec:our_approach}
In this section, we describe our false negative detector (FND) that identifies the failure of a deployed traffic sign detector (TSD). We assume that the weight of the TSD model will be fixed during the deployment phase. 

The proposed FND works in two steps as follows:
\begin{itemize}
\item Collect features from specific areas of an input image, where TSD has not detected any sign.
\item Evaluate those features to identify false negative traffic signs from those areas.
\end{itemize}

The FND relies on the observation that when the TSD misses a sign, most of the time, there are still some excited regions in its internal feature maps, some of which correspond to the location of that sign, see Figure~\ref{fig:main_idea}. We will exploit this fact to build a classifier that takes features from those regions and determine if TSD has failed to detect a sign in that area or not.

Figure~\ref{fig:examples} shows an example of a missed sign with multiple excited regions in the feature maps of the detector. TSD has missed to detect a traffic sign for the input image ($I$). However, the figure also shows one of the internal feature maps corresponding to $I$. We can see multiple excited regions ($\mathcal{R}$) in this feature map. One of these regions are located at the corresponding position of the missing traffic sign. We refer this region as a \textbf{failure} because TSD has failed to detect a sign from here. Other regions will be referred as \textbf{imposter} because those are excited but not related to the missing traffic sign. After binarizing the feature map (step b) in Figure~\ref{fig:examples}), we apply contour area detection to locate the bounding box ($[x_{min}, y_{min}, x_{max}, y_{max}]$) for each excited region ($R_i$). Step d in Figure~\ref{fig:examples} is the output of the FND showing the discovered false negative traffic sign.

\subsection{Training the false negative detection system}
During the training stage, we convert the excited regions coordinates from feature space to image space and measure the intersection over union with the ground truth bounding boxes. Then we label each region using Equation \ref{eqn:failure_imposter_labelling}.

\begin{equation}
Label(R_i)=\begin{cases}
			failure, & G(R_i) \geq \gamma \\
            imposter, & \text{otherwise}
		 \end{cases}
\label{eqn:failure_imposter_labelling}    
\end{equation}
where $G(R_i)$ measures the maximum intersection over union for region $R_i$ with all the ground truths of the input image ($I$). 

Now we have a set of failure and imposter regions, we extract corresponding failure and imposter features from all of the regions ($\mathcal{R}$). To do so, we first stack $N$ three dimensional features maps ($W \times H \times C_i$) from the deployed TSD network along their channel axis, $C$. Each feature map has variable number of channels. 

After the stacking, we get a new feature map ($\mathcal{V}$) of size $W \times H \times K$ where $K = C_1 + \ldots + C_N$. We apply Algorithm \ref{alg:algorithm2} to extract $K$ length 1-dimensional feature for each region ($R_i$) from $\mathcal{V}$.
\begin{figure}[t]
    \centering
    \includegraphics[width=0.9\linewidth]{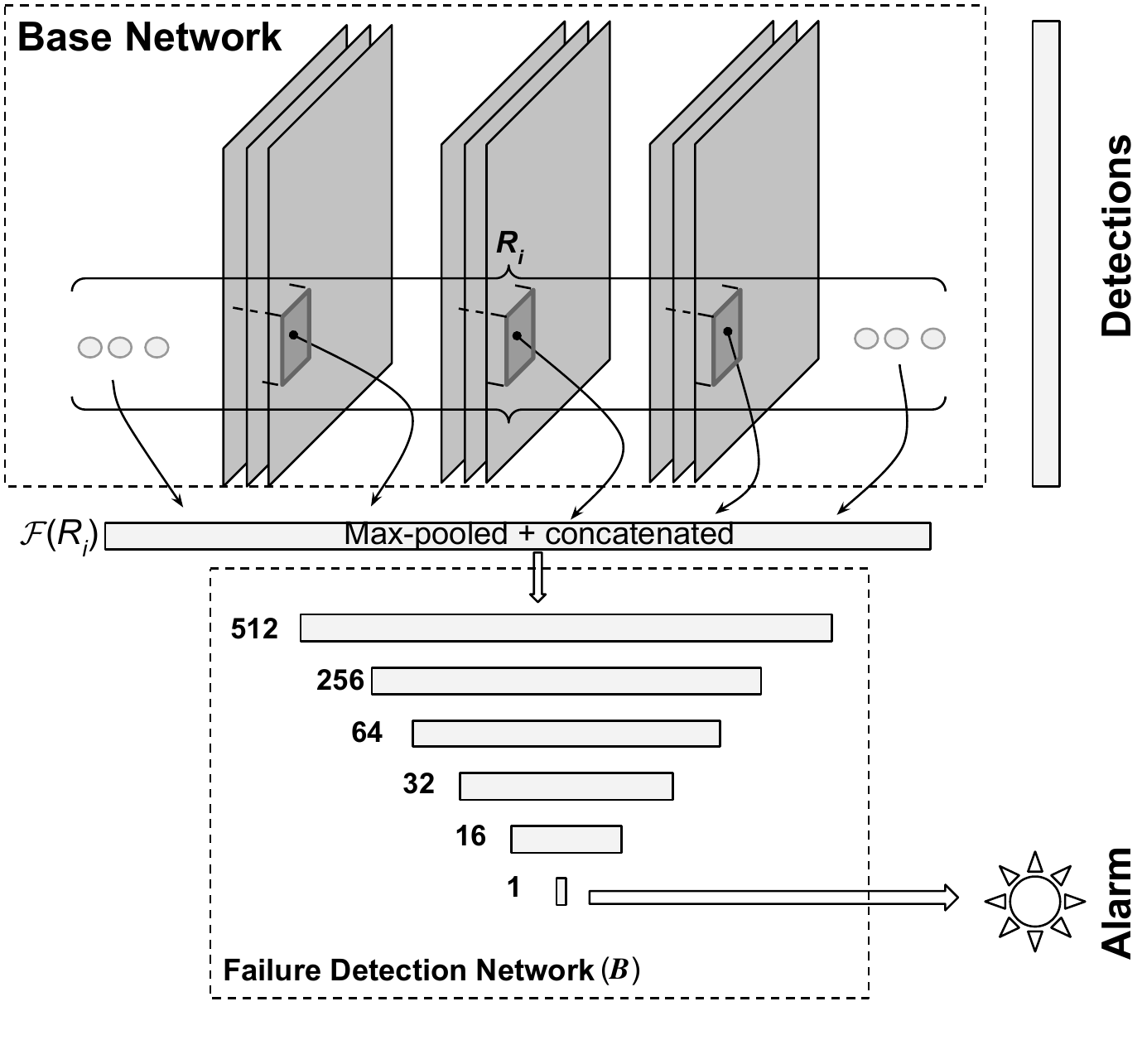}
    \caption{An outline of the proposed false negative detection system. Although the base network has failed to detect a sign, some of its internal feature maps are excited in the corresponding location of that missing sign. The feature extraction pipeline (Figure~\ref{fig:examples}) extracts features from those regions. During deployment, the failure detection network uses these features to discover false negatives traffic signs.}
    \label{fig:Network_Archi}
\end{figure}

\begin{algorithm}[b]
\caption{Feature Extraction Algorithm}
\begin{algorithmic}[1]
\Require $\mathcal{R}$, $\mathcal{V}$
\Ensure $\mathcal{F}$
\ForEach{$R_i \in \mathcal{R}$}
    \State $\mathcal{F}(R_i) = \bigcup\limits_{k=1}^{K}f(\mathcal{V}_k, R_i)$
\EndFor
\end{algorithmic}
\label{alg:algorithm2}
\end{algorithm}

In Algorithm \ref{alg:algorithm2}, $f(\mathcal{V}_k, R_i)$ returns a feature vector as the maximum values within the region ($R_i$) along the $k^{th}$ channels of $\mathcal{V}$.

After extracting $K$ length failure and imposter features vectors, we train a fully connected binary classifier ($\mathcal{B}$) to classify these two types of features. The full architecture of the proposed system is shown in Figure~\ref{fig:Network_Archi}.

\subsection{Deployment of the false negative detection system}
During the testing phase, FND follows the feature extraction pipeline (Figure~\ref{fig:examples}) to extract features from the internal layers of TSD. At first, FND receives the detection output generated by TSD and locate the input image area without any detection. Excited regions are located from this area following the similar approach of step b and step c of the pipeline. These regions are then used to extract features from the internal layers. At the next step, the failure detection network ($\mathcal{B}$) predicts these features as failures or imposters.

\section{Experimental Setup}
\label{sec:experiments}
In this section we describe the datasets, the traffic sign detector and the evaluation metrics that we use to evaluate our proposed method.

\subsection{Datasets}
Our training and testing dataset consist of images from Belgium Traffic Sign Dataset (BTSD)~\cite{timofte2014multi} and German Traffic Sign Detection Benchmark (GTSDB)~\cite{Houben-IJCNN-2013}. The training split of BTSD has been used for all of the training purposes. The testing split of BTSD and the whole dataset from GTSDB has been used only for evaluation purposes (i.e., neither the sign detector nor the false negative detector has seen the images from BTSD testing split and GTSDB during training). 

Rain and fog effect has been applied using Automold \cite{automold} on our test split to simulate different environments in our test dataset. 

There are three groups of signs in both datasets. These are mandatory, prohibitory and danger \cite{Houben-IJCNN-2013}. Table \ref{tbl:images_signs_number} shows the number of images and sign classes in our training and testing settings. 

\begin{table}[bh]
\caption{Number of images and traffic signs of each class.}
\centering
\begin{tabular}{@{}llll@{}}
\cmidrule(l){2-4}
                                  & BTSD (train) & BTSD (test) & GTSDB (test) \\ \cmidrule(l){2-4} 
\multicolumn{1}{l|}{Prohibitory}  & 1694         & 914           & 360     \\
\multicolumn{1}{l|}{Mandatory}    & 996          & 729           & 141     \\
\multicolumn{1}{l|}{Danger}       & 973          & 519           & 99     \\
\multicolumn{1}{l|}{Total  sign}  & 3663         & 2162           & 600     \\
\multicolumn{1}{l|}{Total  image} & 3383         & 1850           & 442     \\ \bottomrule
\end{tabular}
\label{tbl:images_signs_number}
\end{table}

\subsection{Evaluation Metrics}
In this paper, we are using precision and recall to evaluate the proposed method. For false negative detection, precision and recall is defined as follows:
\begin{equation}
    precision = \frac{t_f}{t_f+f_a}
    \label{eqn:fnp}
\end{equation}
where, $t_f$ (true failure) denotes the number of cases where the FND has successfully discovered a failure by the TSD. $f_a$ (false alarm) is a case where FND has mistakenly identified an imposter instance as a failure.

\begin{equation}
    recall = \frac{t_f}{t_f+f_i}
    \label{eqn:fnr}
\end{equation}
where, $f_i$ is the number of failure instances that have been identified as imposter.

\subsection{Traffic Sign Detector}
In all of our experiments, we have used Single Shot Multi-box Detector (SSD) \cite{liu2016ssd} for traffic sign detection. To train the detector for traffic sign, we use SSD with Inception~V2 \cite{szegedy2016rethinking} pre-trained on COCO dataset \cite{lin2014microsoft} from Tensorflow object detection API \cite{huang2017speed}. The minimum score threshold $(\lambda$) is set to $0.5$ for all of the experiments. $\lambda$ is used to filter out traffic signs from TSD generated proposals. Table~\ref{tbl:tsd_ap} shows the detection performance of TSD on the BTSD and GTSDB testing data.

\begin{table}[h]
\centering
\caption{TSD Average precision at 0.5 IOU for BTSD and GTSDB testing data}
\begin{tabular}{@{}lccc@{}}
\cmidrule(l){2-4}
                                  & Prohibitory & Mandatory & Danger \\ \cmidrule(l){2-4} 
\multicolumn{1}{l|}{BTSD (test)}  & 0.86        & 0.87      & 0.90   \\
\multicolumn{1}{l|}{GTSDB (test)} & 0.74        & 0.93      & 0.73   \\ \bottomrule
\end{tabular}
\label{tbl:tsd_ap}
\end{table}

\subsection{Baselines}
In this section, two baseline approaches are proposed to compare the performance of FND.

\textbf{Baseline 1}: To train this baseline, all the traffic signs are cropped from BTSD training data and grouped according to their class (prohibitory, mandatory and danger). The next step is to train an imagenet pre-trained VGG16 classifier with dropout layer to classify these three classes. During testing, we use TSD to detect traffic signs from both BTSD and GTSDB testing data. TSD discards some proposals for having a score less than $\lambda$. This baseline uses the classifier to classify all those rejected proposals and measure the classifier uncertainty using dropout sampling. A lower uncertainty means the classifier has detected a traffic sign with high confidence in the rejected proposals.  

\textbf{Baseline 2}: Similar approaches like \cite{zhang2014predicting, daftry2016introspective} is adopted for the baseline 2 training. TSD is used to detect traffic signs from the BTSD training split, and we collect proposals where TSD score is less than $\lambda$. These proposals are divided into two groups. The first group is named \textit{failure} and contains proposals where TSD has made a mistake by not detecting a sign. The second group is named \textit{imposter} and contains proposals where there is no sign. An imagenet pre-trained VGG16 binary classifier is trained using these two groups. During testing, the classifier assigns a failure score from $0$ to $1$ for each input proposals.

\subsection{False Negative Detector (FND)}
To train the FND, failure and imposter features are collected based on the procedures described in Section \ref{sec:our_approach}. In our experiments, all the images from BTSD training split have been used for the training of TSD and FND. 

For feature collection, We have selected all the three dimensional ($64 \times 64 \times K$) convolutional feature maps in the base network of TSD (Inception V2). Here each feature map has similar width and height and different number of channels, and there are $33$ such feature maps in TSD. 
One of the feature maps has been selected empirically to locate the excited regions. A $64 \times 64 \times 4640$ feature map is generated after stacking all of these feature maps along  their channel axis. To label the excited regions, FND uses Equation~\ref{eqn:failure_imposter_labelling} with $\gamma = 0.5$.

During the training phase, Algorithm~\ref{alg:algorithm2} is used to collect features for each region. These features are used to train a binary classifier (Figure~\ref{fig:Network_Archi}) to detect failure and imposter features.

\section{Evaluation and Results}\label{sec:eval}
\subsection{Naive Solution}

Before we present the results of our proposed method, we will discuss why the simple act of lowering the threshold to accept more detections by the sign detector does not provide a satisfactory solution. 

\begin{figure}[t]
\centering
\begin{subfigure}{0.50\columnwidth}
\centering
    \includegraphics[width=\columnwidth]{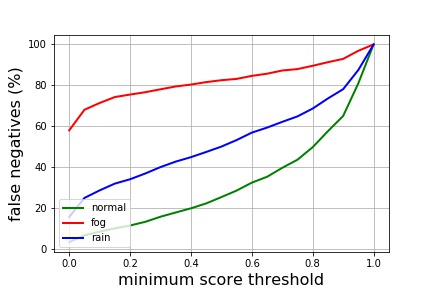}
    \caption{}
    \label{fig:belgium_false_negatives}
\end{subfigure}%
\begin{subfigure}{0.50\columnwidth}
\centering
    \includegraphics[width=\columnwidth]{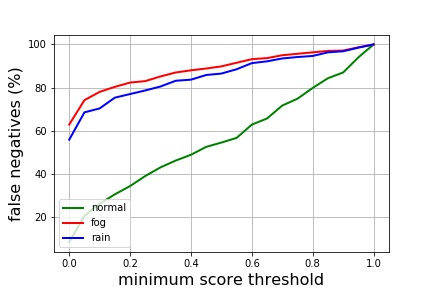}
    \caption{}
    \label{fig:germany_false_negatives}
\end{subfigure}
\label{fig:belfium_germany_false_negatives}
\caption{The false negatives rate in BTSD and GTSDB for different minimum score threshold.}
\end{figure}

Although lowering the minimum score threshold decreases the number of false negatives, this threshold needs to be tuned for different operational condition and environment. Figure~\ref{fig:belgium_false_negatives} and Figure~\ref{fig:germany_false_negatives} show the percentage of false negatives generated by TSD for three different settings (normal, fog and rain) of BTSD and GTSDB dataset. 

In Figure \ref{fig:belgium_false_negatives}, the percentage of false negatives for $\lambda = 0.5$ in normal BTSD test dataset is $25\%$. However, it increases to $50\%$ and $60\%$ respectively for rain and fog condition. To maintain similar false negative rate as in the normal environment, for both fog and rain, we need to accept all of the TSD generated proposals. 

Figure \ref{fig:germany_false_negatives} shows the TSD generated false negative when TSD is trained using BTSD and tested in GTSDB. We can also see the higher false negative rate for fog and rain than normal weather condition. Besides, $\lambda$ related to BTSD dataset will not work here. This experiment shows the necessity of a separate false negative detection system rather than tuning the minimum score threshold of TSD. 

\subsection{Comparison to baselines}
We have tested baseline 1, baseline 2 and FND on the normal, foggy and rainy version of BTSD and GTSDB testing data. The purpose of this different settings is to evaluate the robustness of the proposed method in variable conditions. 

\begin{figure}[t]
\centering
\begin{subfigure}{0.50\columnwidth}
\centering
    \includegraphics[width=\columnwidth]{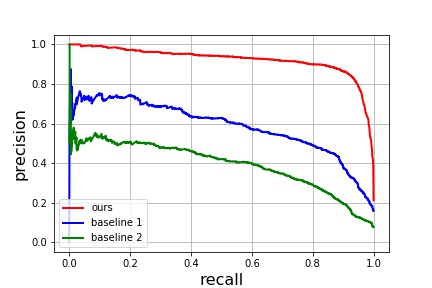}
    \caption{}
    \label{fig:normal_belgium}
\end{subfigure}%
\begin{subfigure}{0.50\columnwidth}
\centering
    \includegraphics[width=\columnwidth]{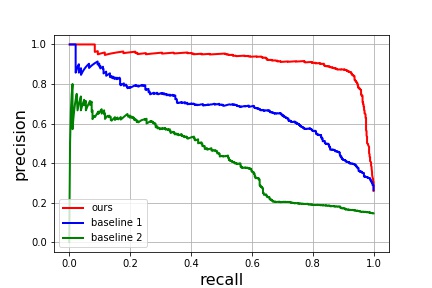}
    \caption{}
    \label{fig:normal_germany}
\end{subfigure}
\label{fig:normal_belgium_germany}
\caption{Precision vs Recall curve for (a) Belgium and (b) German test data in normal weather condition.}
\end{figure}

Figure \ref{fig:normal_belgium} shows the comparison of precision and recall curve for normal BTSD testing data. For $80\%$ recall FND achieves $89.97\%$ precision. For similar recall, precision for baseline 1 and baseline 2 are $28.87\%$ and $48.96\%$ respectively. 

\begin{figure}[t]
\centering
\begin{subfigure}{0.50\columnwidth}
\centering
    \includegraphics[width=\columnwidth]{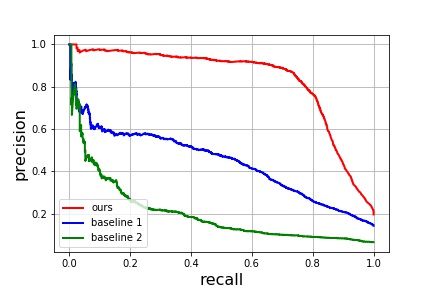}
    \caption{}
    \label{fig:foggy_belgium}
\end{subfigure}%
\begin{subfigure}{0.50\columnwidth}
\centering
    \includegraphics[width=\columnwidth]{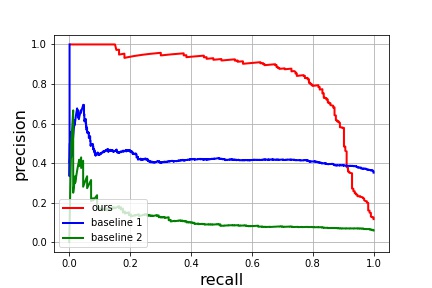}
    \caption{}
    \label{fig:foggy_germany}
\end{subfigure}
\label{fig:foggy_belgium_germany}
\caption{Precision vs Recall curve for (a) Belgium and (b) German test data in foggy weather condition.}
\end{figure}

\begin{figure}[t]
\centering
\begin{subfigure}{0.50\columnwidth}
\centering
    \includegraphics[width=\columnwidth]{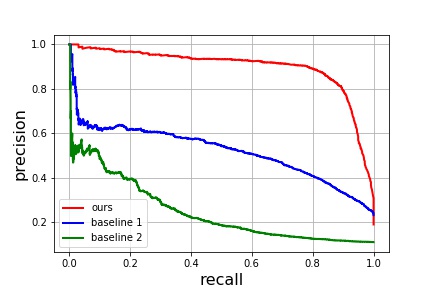}
    \caption{}
    \label{fig:rainy_belgium}
\end{subfigure}%
\begin{subfigure}{0.50\columnwidth}
\centering
    \includegraphics[width=\columnwidth]{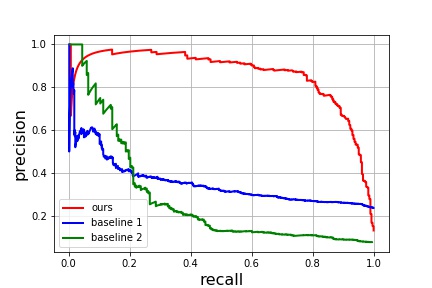}
    \caption{}
    \label{fig:rainy_germany}
\end{subfigure}
\label{fig:rainy_belgium_germany}
\caption{Precision vs Recall curve for (a) Belgium and (b) German test data in rainy weather condition.}
\end{figure}

Figure \ref{fig:foggy_belgium} and Figure \ref{fig:rainy_belgium} show the performance of FND, baseline 1 and baseline 2 in two different weather conditions of BTSD testing data. For simulated rainy weather FND achieves $89.07\%$ precision and it becomes $76.5\%$ for foggy weather. Precision for baseline 1 drops from $28.8\%$ to $12.7\%$ for rainy weather and $9.0\%$ for foggy weather. Baseline 2 also suffers from changing environment.  Precision drops from $48.9\%$ to $40.8\%$ for rainy data and $26.0\%$ for foggy data.

In GTSDB, for normal dataset, FND achieves $90.8\%$ precision for $80.0\%$ recall. Figure \ref{fig:normal_germany} shows the precision recall curve for FND, baseline 1 and baseline 2 in GTSDB for normal weather condition. Though FND has been trained using data on BTSD, it can raise alarm with a high precision and recall when the traffic sign detector makes a mistake. This high alarm rate proves the robustness of the collected features for failure and imposter classification. 

We have also tested FND on simulated rain and fog version of GTSDB. For $80.0\%$ recall, FND achieves $82.98\%$ and $79.30\%$ precision respectively. Figure \ref{fig:rainy_germany} and Figure \ref{fig:foggy_germany} shows the precision and recall curve for FND, baseline 1 and baseline 2 in rain and fog version of GTSDB.

\begin{figure*}[t]
    \centering
    \includegraphics[width=\textwidth]{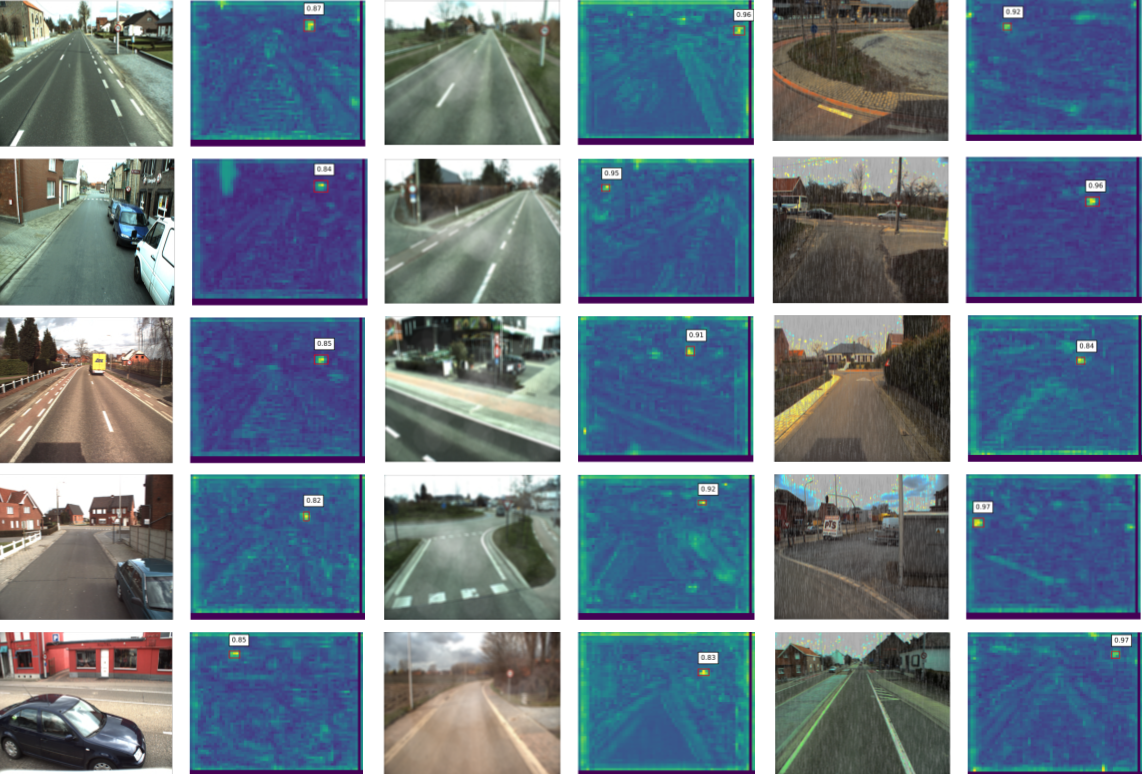}
    \caption{Examples of false negative detection by FND. The 1\ts{st}, 3\ts{rd} and 5\ts{th} columns show input images for TSD from normal, foggy and rainy weather respectively. In all of these cases, TSD has failed to detect the traffic sign presented in the input. The 2\ts{nd}, 4\ts{th} and 5\ts{th} column show the failure prediction score generated by FND for the false negative traffic sign.}
    \label{fig:all_samples}
\end{figure*}

Figure~\ref{fig:all_samples} shows a qualitative results of our false negative detection system. Here we show some cases where FND has successfully identified false negative traffic signs missed by the detector. These sample results are taken from three different environments (normal, simulated fog and simulated rain) of BTSD testing data.

\section{Conclusions and Future Work}\label{sec:conclusions}
In this paper, we addressed the critical aspect of detecting false negatives of object detectors in the context of traffic sign detection for autonomous vehicles. This is an important safety issue. If the vehicle fails to locate any critical sign, it might make a catastrophic failure. We proposed a false negative detector (FND) that is trained to distinguish missed signs from imposters in the excitations of the feature maps of a deployed traffic sign detector.  We tested FND using two traffic sings benchmarking datasets, the Belgium Traffic Sign Detection dataset and German Traffic Sign Recognition Benchmark dataset, as well as simulated weather conditions, fog and rain using images from both datasets. We compared our proposed method to two baselines and showed that it provides better performance in detecting false negatives. Future work will focus on identifying the best layers in TSD to extract more effective excited regions for the false negative detector.

\bibliographystyle{IEEEtran}
\bibliography{bib}
\end{document}